\documentclass[a4paper,fleqn]{cas-dc}

\usepackage[numbers]{natbib}
\usepackage{graphicx}
\usepackage{amsmath,amssymb,amsfonts}
\usepackage{algorithmic}
\usepackage{textcomp}
\usepackage{subfigure} 
\newcommand{\etal}{\textit{et al.}\@ }
\newcommand{\ie}{\textit{i.e.}\@ }
\usepackage{multirow}

\def\tsc#1{\csdef{#1}{\textsc{\lowercase{#1}}\xspace}}
\tsc{WGM}
\tsc{QE}

\begin{document}
\let\WriteBookmarks\relax
\def\floatpagepagefraction{1}
\def\textpagefraction{.001}

\shorttitle{Improving GANs with a Feature Cycling Generator}    

\shortauthors{S. Park and Y.-G. Shin}  

\title [mode = title]{Improving GANs with a Feature Cycling Generator}  



%

\author[1]{Seung Park}[type=editor]

\ead{spark.cbnuh@gmail.com}

\affiliation[1]{organization={Biomedical Engineering, Chungbuk National University Hospital},
            addressline={776}, 
            city={Seowon-gu, Cheongju-si},
            state={Chungcheongbuk-do},
            country={Rep. of Korea}}

\author[2]{Yong-Goo Shin}[type=editor]
\cormark[1]

\ead{ygshin92@korea.ac.kr}

\affiliation[2]{organization={Department of Electronics and Information Engineering, Korea University},
            addressline={2511, Sejong-ro, Jochiwon-eup}, 
            city={Sejong-si},
            postcode={30019}, 
            country={Rep. of Korea}}

\cortext[1]{Corresponding author}


\begin{abstract}
Generative adversarial networks (GANs), built with a generator and discriminator, significantly have advanced image generation. Typically, existing papers build their generators by stacking up multiple residual blocks since it makes ease the training of generators. However, some recent papers commented on the limitation of the residual block and proposed a new architectural unit that improves the GANs performance. Following this trend, this paper presents a novel unit, called feature cycling block (FCB), which achieves impressive results in the image generation task. Specifically, the FCB has two branches: one is a memory branch and the other is an image branch. The memory branch keeps meaningful information at each stage of the generator, whereas the image branch takes some useful features from the memory branch to produce a high-quality image. To show the capability of the proposed method, we conducted extensive experiments using various datasets including CIFAR-10, CIFAR-100, FFHQ, AFHQ, and subsets of LSUN. Experimental results demonstrate the substantial superiority of our approach over the baseline without incurring any objective functions or training skills. For instance, the proposed method improves Frechet inception distance (FID) of StyleGAN2 from 4.89 to 3.72 on the FFHQ dataset and from 6.64 to 5.57 on the LSUN Bed dataset. We believe that the pioneering attempt presented in this paper could inspire the community with better-designed generator architecture and with training objectives or skills compatible with the proposed method.
\end{abstract}



\maketitle

\section{Introduction}
\label{section1}
Generative adversarial networks (GANs)~\cite{goodfellow2014generative}, which consist of a generator and a discriminator, have significantly advanced in various research fields such as image generation~\cite{karras2019style, karras2021alias, miyato2018spectral, zhao2020improved}, text-to-image translation~\cite{hong2018inferring, reed2016generative}, image-to-image translation~\cite{isola2017image, choi2018stargan, zhu2017unpaired}, and image inpainting~\cite{yu2018free, sagong2019pepsi, shin2020pepsi++}. In general, the generator and discriminator compete against each other during training:
the generator aims at emulating the observed image as much as possible in order to fool the discriminator, while the discriminator attempts to distinguish fake samples from real ones. Despite the great effort of training GANs stably from the discriminator side~\cite{wu2021gradient, miyato2018spectral, gulrajani2017improved, kodali2017convergence, zhang2019consistency, zhao2020improved, ross2018improving, mescheder2018training, wu2019generalization, wei2018improving, roth2017stabilizing, yeo2021simple}, such as objective functions and regularization techniques, it has been relatively less explored on the generator architecture. In fact, the competition between the generator and the discriminator is not quite fair~\cite{bai2022glead}. Specifically, in adversarial learning, the discriminator directly accesses the real image, examines how close the real and generated images are, and computes a probability to train itself and affect the generator~\cite{yang2022improving}. Conversely, the generator only can receive the learning signal, \ie gradients, derived from the discriminator. Thus, this unfair competition allows the discriminator to easily win the game, and massive experimental results support evidence for this analysis. Therefore, designing an appropriate generator architecture for fair competition is essential for GANs. 

The generator in a GAN is typically built by stacking up multiple residual blocks~\cite{miyato2018spectral, miyato2018cgans, brock2018large, yeo2022image, karras2020analyzing, karras2019style, park2021GRB, park2022generative} since the shortcut in the residual block well assists the flow of information from the latent space (initial stage) to the image (last stage). However, some recent papers~\cite{park2021GRB, park2022effective, park2021novel, wang2021up} drew attention to the limitations of the residual block and introduced novel modules that are more suitable for building the generator. In detail, Wang~\etal~\cite{wang2021up} proposed an altered residual block that has different convolution kernel sizes from the traditional one and demonstrated that their block shows better performance than the traditional one. However, a recent paper~\cite{park2021GRB} pointed out that the performance improvement in~\cite{wang2021up} is caused by the activation function (gated linear unit) rather than the module architecture. Park~\etal~\cite{park2021GRB} redesigned the residual block by adding a side residual path that effectively emphasizes the informative feature while suppressing the less useful one. In addition, Park~\etal~\cite{park2021novel} introduced novel generator architecture, called ABGAN, which contains an auxiliary branch controlling the information flow between residual blocks. These approaches show impressive performance and imply the possibility of performance improvement via newly designing the generator architecture.

\begin{figure}
\centering
\includegraphics[width=0.8\linewidth]{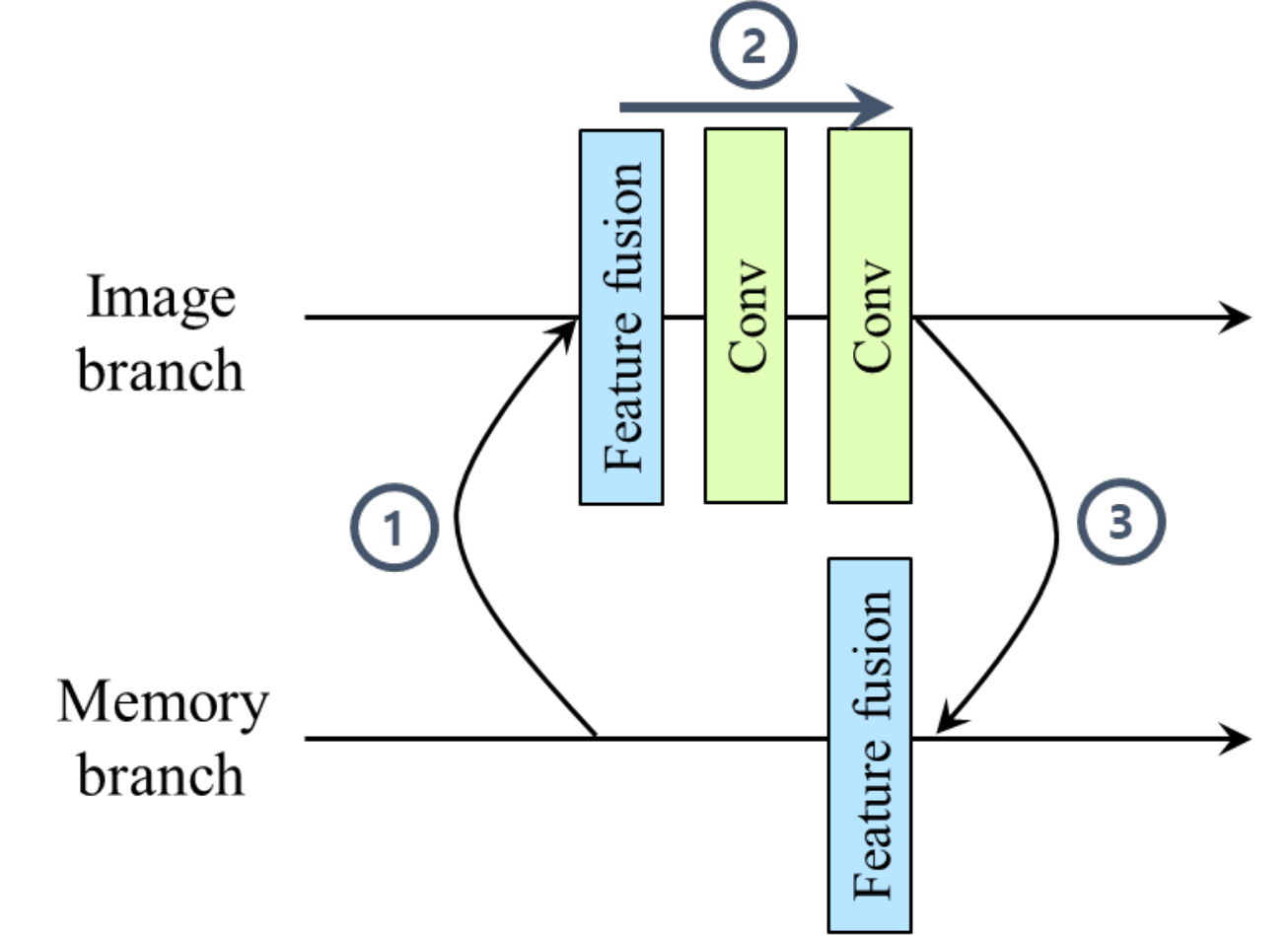}
\caption{Concept diagram of the proposed FCB. The FCB is complementary to the original residual block by appending the memory branch and the feature fusion modules.}
\label{fig1}
\vspace{-0cm}
\end{figure}

Based on these observations, this work investigates a different generator architecture to improve the GAN performance. In particular, this paper offers a novel module used for the generator, called feature cycling block (FCB), where the concept diagram is depicted in Fig.~\ref{fig1}. The FCB contains two branches: the image branch aims to produce features used for producing high-quality images, whereas the memory branch gives selected information to the image branch as well as memorizes meaningful features in the image branch. More specifically, in the FCB, the image branch first takes some useful features from the memory branch through the feature fusion module and then produces features using multiple convolution layers. After that, the memory branch updates its feature using the newly produced features in the image branch. That means the features in both branches are jointly updated by interchanging their information. In order to prove the superiority of the proposed method, we conduct a comprehensive empirical study by evaluating the GAN performance on various datasets including CIFAR-10~\cite{krizhevsky2009learning}, CFIAR-100~\cite{krizhevsky2009learning}, FFHQ~\cite{karras2019style}, AFHQ~\cite{choi2020stargan}, and subsets of LSUN~\cite{yu15lsun}. Quantitative evaluations show that the proposed method can improve the image generation performance, in terms of Frechet inception distance (FID)~\cite{heusel2017gans} and Precision/Recall~\cite{kynkaanniemi2019improved, sajjadi2018assessing}. In the remainder of this paper, we explain the preliminaries in Section~\ref{section2} and introduce the proposed method in Section~\ref{sec3}. In Section~\ref{sec4}, extensive experimental results are presented to show the superior performance of the proposed method compared to conventional ones, and Section~\ref{sec5} concludes the paper. 

We summarize the contributions of this work as follows: 
\begin{itemize}
\item This paper presents a novel module for the generator called feature cycling block (FCB), which achieves superior performance over conventional methods. 
\item We newly design a feature fusion module that successfully blends features from two different sources. 
\item To demonstrate that the FCB benefits the GANs performance, we conduct extensive experiments on various datasets and reveal that the proposed method achieves remarkable performance with various quantitative evaluation metrics. 

\end{itemize}

\section{Related Works}
\label{section2}
\subsection{Generative Adversarial Networks}
\label{sec2.1}
In the training procedure of GANs~\cite{goodfellow2014generative}, the generator \textit{G} is trained to produce samples following the real data distribution $P_{data}(x)$ and the discriminator \textit{D} is optimized to classify the generated real and generated samples. The objective functions of \textit{D} and \textit{G} are defined as

\begin{eqnarray}
\label{eq1}
      \lefteqn{L_D = - E_{x\sim P_\textrm{data}(x)}[\log D(x)]}\nonumber\\
    & & {\qquad \qquad} - E_{z\sim P_{z}(z)}[\log(1-D(G(z)))],
\end{eqnarray}
\begin{eqnarray}
\label{eq2}
    L_G = -E_{z\sim P_{z}(z)}[\log(D(G(z)))],
\end{eqnarray}
where \textit{x} indicates a sample from $P_{data}(x)$ and \textit{z} means a latent vector sampled from latent space $P_z(z)$ such as Gaussian normal distribution. Ideally, minimizing these formulas will arrive at an optimal solution where \textit{G} can generate the entire real image and \textit{D} can no longer differentiate the image source~\cite{goodfellow2014generative}. Despite the ongoing research, however, the competition between \textit{G} and \textit{D} is still unstable since that competition is held in the high-dimensional feature space. Therefore, it is hard to observe the aforementioned ideal case where \textit{G} synthesizes the target real samples perfectly. 

To alleviate this problem, extensive studies have been carried out. Some papers~\cite{wu2021gradient, miyato2018spectral, gulrajani2017improved, kodali2017convergence, zhang2019consistency} noticed that the sharp gradient space in \textit{D} makes GANs training unstable and proposed various regularization~\cite{zhang2019consistency, zhao2020improved, mescheder2018training, ross2018improving} and normalization techniques~\cite{miyato2018spectral, lin2021spectral}. More specifically, existing papers~\cite{gulrajani2017improved, wu2019generalization, wei2018improving, kodali2017convergence, roth2017stabilizing} usually added the regularization term, such as the gradient penalty~\cite{gulrajani2017improved, wu2019generalization, wei2018improving} and $R_{1}$ regularization~\cite{mescheder2018training, karras2020analyzing, karras2019style, karras2021alias}, into their objective functions, which constrains the magnitude of the gradient space. On the other hand, some studies investigated the normalization strategy to constrain the Lipschitz constant in \textit{D} around one. The spectral normalization (SN)~\cite{miyato2018spectral} is the most popular normalization technique and is still widely used in recent studies~\cite{yeo2021simple, yeo2022image, park2022effective}. These regularization or normalization techniques provide an appealing synthesis result, but the instability problem still remains. 

In order to bring performance gain, some papers developed novel modules which can be easily applied to the existing \textit{D} or \textit{G} architectures~\cite{miyato2018cgans, yeo2021simple, park2021GRB, park2022effective, yeo2022image, park2021novel, park2022pconv, sagong2022conditional, park2022generative, karras2020analyzing}. Miyato~\etal~\cite{miyato2018cgans} introduced a conditional projection technique that effectively provides conditional information at the end of \textit{D}. Yeo~\etal~\cite{yeo2021simple} proposed a cascading rejection module that iteratively predicts the source of the input image using orthogonal features. Park~\etal~\cite{park2022pconv} introduced a perturbed convolution (PConv), which is designed to prevent the overfitting problem of \textit{D}. Sagong~\etal~\cite{sagong2022conditional} developed conditional convolution, named cConv, which alters the convolutional kernel according to the given condition. Recently, Park~\etal~\cite{park2022generative} proposed a latent-adaptive convolution technique called GConv, which varies the convolutional kernel according to the input latent vector. These methods simply added their modules to the baseline model without modifying the baseline model. Differently, this paper rebuilds the \textit{G} architecture by stacking up the FCB. 

\begin{figure*}
\centering
\includegraphics[width=0.95\linewidth]{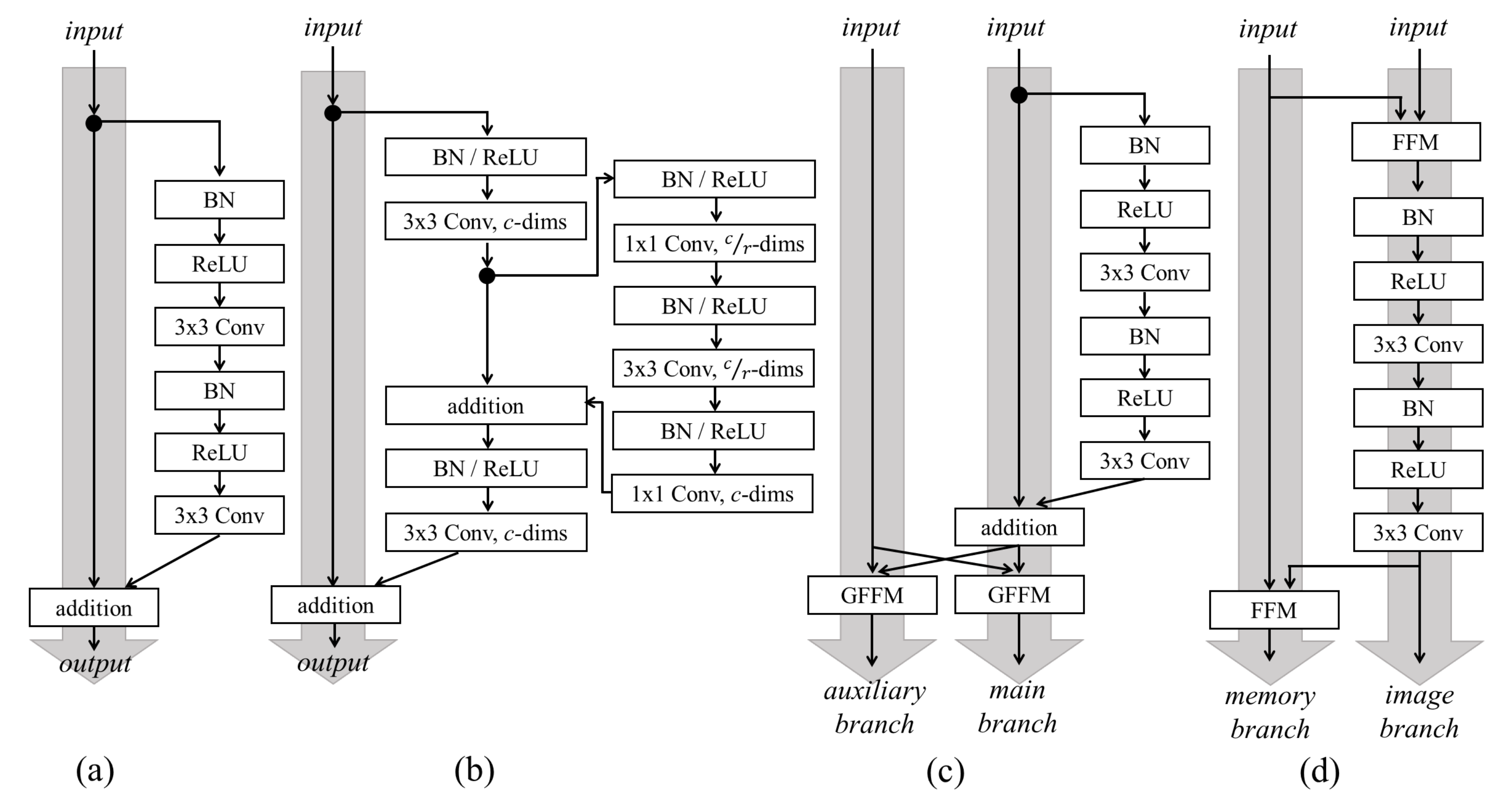}
\caption{Comparison of the block modules for \textit{G}. (a) The traditional residual block, (b) the residual block with a side residual branch~\cite{park2022generative}, (c) the residual block with an auxiliary branch
~\cite{park2021novel}. GFFM indicates the gated feature fusion module. (d) the proposed feature cycling block (FCB). FFM indicates the feature fusion module.}
\label{fig2}
\vspace{-0cm}
\end{figure*}

\subsection{Conventional block modules for \textit{G}}
\label{sec2.2}
As mentioned in Sec.~\ref{section1}, recent studies build their \textit{G} using multiple residual blocks~\cite{miyato2018spectral, miyato2018cgans, brock2018large, yeo2022image, karras2020analyzing, karras2019style, park2021GRB, park2022generative}, where the residual block is described in Fig~\ref{fig2}.(a). We agree that users can use the modulated weight convolution~\cite{karras2020analyzing} instead of batch normalization (BN) or other activation functions such as Leaky-ReLU and gated linear unit; we just draw the common residual block. In~\cite{park2021GRB}, as shown in Fig.~\ref{fig2}(b), they added a side residual path containing a bottleneck path that emphasizes the informative feature while suppressing the less useful one. This block requires a little additional computational cost due to the side path but shows better performance than the traditional residual block. Fig.~\ref{fig2}(c) describes the architectural module used in ABGAN~\cite{park2021novel}. In ABGAN, they added the auxiliary branch to the side of the traditional residual block, where the main branch produces the image feature via the residual block and the auxiliary branch directly conveys the coarse information in the earlier layer to the later one. This method exhibits remarkable performance and implies the possibility of performance gain by newly designing the block module for \textit{G}. Inspired by these approaches, we attempt to design a new block module that is suitable for the modern baseline network such as StyleGAN2. 

\section{Proposed Method}
\label{sec3}
Fig.~\ref{fig2}(d) shows the overall architecture of the proposed method, which consists of multiple convolutional layers and feature fusion modules (FFMs). In the remainder of this section, we will introduce how the FCB works first, and then explain the FFM used at the FCB. 
 
\subsection{Feature cycling block (FCB)}
\label{sec3.1}
The FCB consists of two different branches: the memory branch and the image branch (left and right branch in Fig.~\ref{fig2}(d), respectively). The memory branch takes the features projected from the latent space as input, whereas the image branch uses trainable constant values. That means the memory branch aims to keep meaningful information obtained from the latent space, \ie latent features, and to transfer it to the image branch. In contrast, the goal of the image branch is to produce image features with the help of the memory branch. To achieve these goals, we design the FCB in which both branches alternately send and receive information. Specifically, the memory branch first provides latent features to the image branch via the FFM that effectively blends the features from the two different branches; the image branch gets the latent features. (A detailed FFM architecture will be explained in the next subsection.) Then, the image branch updates the blended features using multiple convolution, normalization, and activation operations like the traditional residual block in Fig.~\ref{fig2}(a). It is worth noting that the user can modulate the convolution types or operation orders in the image branch. For example, when the user decides the baseline model as StlyeGAN2~\cite{karras2020analyzing}, there is no matter even using the modulated weight convolution instead of BN and changing the order as FFM-Conv-LeakyReLU-Convs-LeakyReLU. After that, using the output features of the image branch, the latent features in the memory branch are updated through another FFM located in the memory branch. 

Here, there are two major differences between the FCB and conventional methods. One is that, unlike the typical residual block, the FCB contains an additional path, \ie, the memory branch, which supports better information transfer than the shortcut in the residual block. As proved in~\cite{park2021novel}, this additional path greatly helps \textit{G} generate high-quality images. The other difference is that, similar to the feedback system, the output of the FCB is circled back and used to update the input of the FCB, \ie latent features in the memory branch; the input latent feature gets feedback to become a more proper feature for the image branch. Indeed, the residual block in Fig.~\ref{fig2}(c) contains the auxiliary path like the FCB, but it simultaneously updates features in the auxiliary and main branches at the end of the block. That means, this method uses the auxiliary branch as an additional shortcut path, and nothing more. Contrary to this approach, the FCB block sequentially updates features in the image and memory branches. This difference might be simple but it greatly improves the representation power of the network rather than the conventional method. We will confirm this argumentation in the empirical experiments with various ablation studies, in Section~\ref{sec4.2}.

\begin{figure}
\centering
\includegraphics[width=1.0\linewidth]{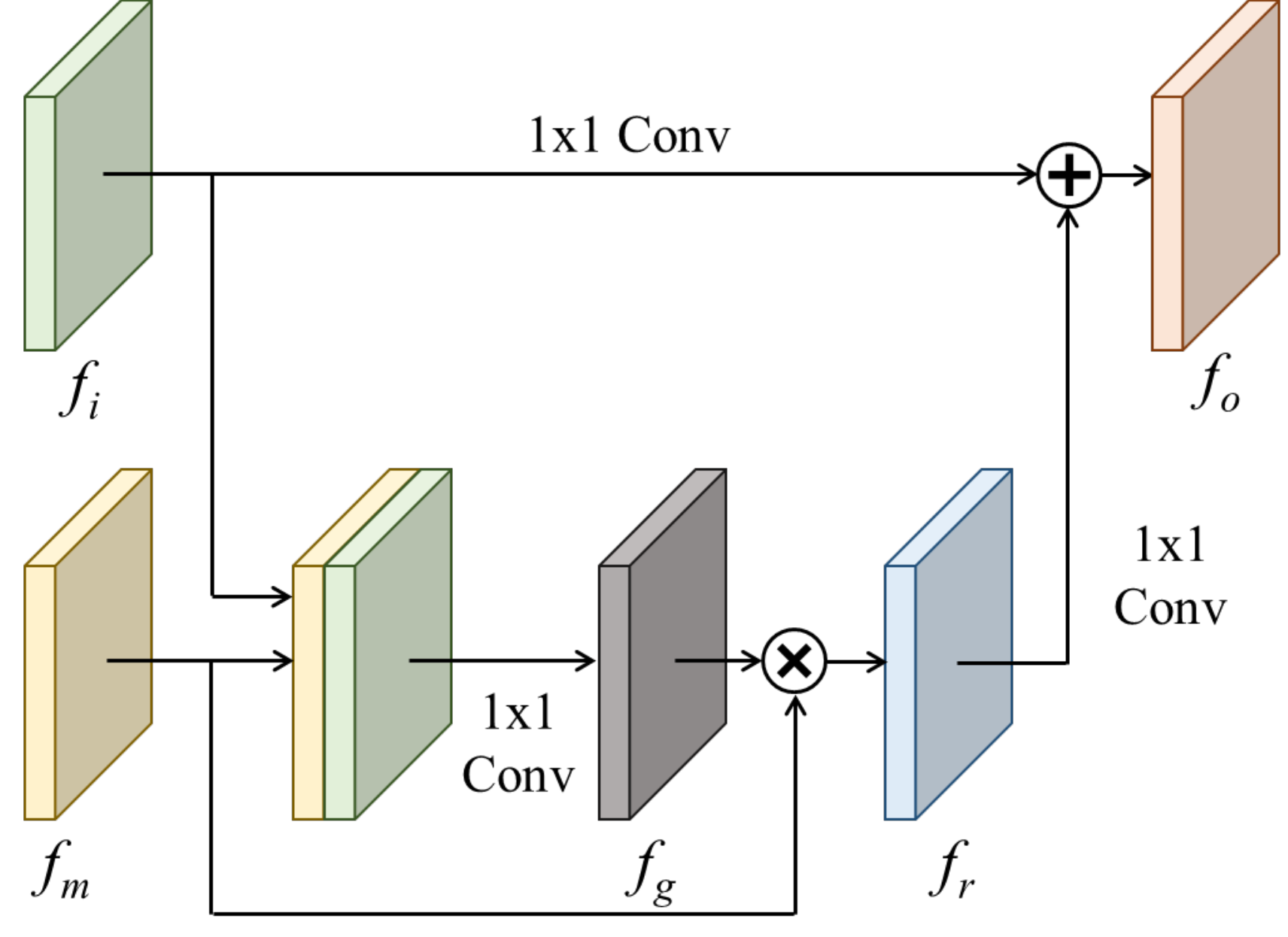}
\caption{The illustration of the feature fusion module (FFM). In the FFM, $f_i$ ($f_m$) is updated by taking meaningful features from $f_m$ ($f_i$). The FFM in this figure is used for updating the feature in the image branch $f_i$. When updating the feature in the memory branch, the locations of $f_i$ and $f_m$ are swapped.}
\label{fig3}
\vspace{-0cm}
\end{figure}

\subsection{Feature fusion module (FFM)}
\label{sec3.2}
To combine features from image and memory branches effectively, we proposed the FFM as depicted in Fig.~\ref{fig3}. Fig.~\ref{fig3} describes the FFM used for updating the feature in the image branch $f_i$, \ie the first (upper) FFM in the FCB. When updating the feature in the memory branch, \ie the FFM in the bottom side of the FCB, the locations of $f_i$ and $f_m$ in Fig.~\ref{fig3} are swapped. In the remainder of this subsection, we will explain how the FFM works when updating the $f_i$. 

In order to extract meaningful features from $f_m$, we used the gating mechanism that selects the features effectively. Toward this end, we first concatenate $f_i$ and $f_m$ and produce the gating map $f_g$ using the $1\times1$ convolutional layer which has a sigmoid as an activation function. In short, $f_g$ is computed as 
\begin{equation}
    f_g = \sigma (Conv(f_i \odot f_m))),
\label{eq3}
\end{equation}
where \textit{Conv} and $\odot$ indicate the $1\times1$ convolutional layer and channel concatenation operation, respectively, whereas $\sigma$ means the sigmoid function. Using $f_g$, we produce the refined features $f_r$ by multiplying $f_m$ and $f_g$. That means, using the $f_g$, we try to retain information useful to $f_i$ and remove the unuseful one. Then, we linearly blend $f_r$ and $f_i$ to produce the output feature $f_o$ as follows:
\begin{equation}
    f_{o} = \frac{1}{\sqrt{2}}(Conv(f_i) + Conv(f_r)). 
\label{eq4}
\end{equation}
The FFM architecture is simple and intuitive but shows impressive performance with the FCB. We will prove the superiority of the FFM in Sec.~\ref{sec4.2}.

\begin{table}[t]
\caption{Network architectures of the generator and discriminator for CIFAR-10 and CIFAR-100 datasets (top) and for FFHQ and LSUN datasets (bottom). We set the network architectures in~\cite{miyato2018spectral} and~\cite{karras2020analyzing} as our baseline models, respectively. Block* means that the conventional or proposed blocks, such as the residual block and proposed FCB, can be located there.}
\centering
\vspace{2mm}

\begin{tabular}{c|c}
\hline\hline
Generator & Discriminator \\
\hline
$z \in \mathbb{R}^{128} \sim N(0, I)$ & RGB image \\
FC, $4 \times 4 \times 256$ & ResBlock, down, 128 \\
Block*, up, 256 & ResBlock, down, 128\\
Block*, up, 256 & ResBlock, 128 \\
Block*, up, 256 & ResBlock, 128 \\
BN, ReLU & ReLU \\
$3\times3$ Conv, Tanh & Global sum pooling \\
& Dense, 1 \\
\hline\hline
\end{tabular}

\vspace{0.5cm}
\begin{tabular}{c|c}
\hline\hline
Generator & Discriminator \\
\hline
(Synthesis network) & RGB image \\
Constant, $4 \times 4 \times 512$ & $1\times1$ Conv, 64 \\
Block*, up, 512 & ResBlock, down, 64\\
Block*, up, 512 & ResBlock, down, 128 \\
Block*, up, 512 & ResBlock, down, 256 \\
Block*, up, 256 & ResBlock, down, 512 \\
Block*, up, 128 & ResBlock, down, 512 \\
Block*, up, 64 & ResBlock, down, 512 \\
$1\times1$ Conv & Mini-batch Std \\
 & Dense, 512 \\
& Dense, 1 \\
\hline\hline
\end{tabular}

\vspace{0.2cm}
\label{table1}
\end{table}

\section{Experiments}
\label{sec4}
\subsection{Implementation Details}
\label{sec4.1}
\noindent
\newline
\textbf{Dataset} In order to show the superiority of the proposed method, we performed extensive experiments with several datasets: CIFAR-10~\cite{krizhevsky2009learning}, CIFAR-100~\cite{krizhevsky2009learning}, FFHQ~\cite{karras2019style}, AFHQ~\cite{choi2020stargan}, and subsets of LSUN~\cite{yu15lsun}. More specifically, the CIFAR-10 and CIFAR-100 datasets consist of 60k images with $32\times32$ resolution, whereas the FFHQ dataset has 70k images with $1024\times1024$ resolution. The AFHQ is a dataset of animal faces consisting of 15,000 high-quality images at $512\times512$ resolution. In our experiments, to train the network on the FFHQ and AFHQ datasets with a single GPU, we resized the resolution of those datasets to $256\times256$. On the other hand, among the various classes in the LSUN dataset, we employed the car and horse images to train the network. Since the LSUN dataset contains numerous images in each class, \textit{e.g.} the car has 5.5M images, we sub-sampled 200k images in each class. We also resized the LSUN images to $256\times256$ resolution. \newline

\begin{table*}[!t]
\caption{Comparison of the GAN performances of the proposed method with those of the conventional methods on the CIFAR-10 and CIFAR-100 datasets in terms of FID.}

\begin{center}
\begin{tabular}{ c | c | c | c | c | c | c}
\hline\hline
\multirow{2}*{Dataset} &  & SNGAN & BigGAN & GRB & ABGAN & FCB \\
 &  & \cite{miyato2018spectral} & \cite{brock2018large} & \cite{park2021GRB} & \cite{park2021novel} & (Proposed) \\

\hline
\multirow{4}*{CIFAR-10} & trial 1 & 12.53 & 11.78 & 10.52 & 10.21 & \textbf{9.02} \\
& trial 2 & 13.03 & 12.65 & 10.52 & 10.36 & \textbf{9.60} \\
& trial 3 & 13.40 & 12.31 & 10.97 & 10.56 & \textbf{9.78} \\
\cline{2-7}
& Mean & 12.99 & 12.25 & 10.67 & 10.38 & \textbf{9.46} \\
                       
\cline{1-7}
\multirow{4}*{CIFAR-100} & trial 1 & 17.12 & 16.30 & 14.42 & 14.36 & \textbf{13.18} \\
& trial 2 & 17.21 & 15.28 & 14.48 & 14.35 & \textbf{13.84} \\
& trial 3 & 17.75 & 16.26 & 15.07 & 14.51 & \textbf{13.61} \\
\cline{2-7}
& Mean & 17.36 & 15.94 & 14.66 & 14.41 & \textbf{13.54} \\

\hline\hline
\end{tabular}
\end{center}
\vspace{-0.2cm}
\label{table2}
\end{table*}

\noindent
\textbf{Training Details} As aforementioned in Sec.~\ref{sec2.1}, by minimizing the Eqs.~\ref{eq1} and~\ref{eq2}, $G$ could synthesize $P_{data}(x)$ successfully. However, due to various issues such as the gradient vanishing problem, the training procedure with Eqs.~\ref{eq1} and~\ref{eq2} is often failed. To alleviate this problem, many studies have been carried out. Mao~\etal~\cite{mao2017least} introduced an objective function based on the least square error (LSGAN) and Arjovsky~\etal~\cite{arjovsky2017wasserstein} used Wasserstein distance to build the loss function (WGAN). The current widely-used objective functions are hinge loss~\cite{brock2018large, yeo2021simple, miyato2018cgans, miyato2018spectral, park2022pconv, sagong2019pepsi, shin2020pepsi++, chen2019self, park2021GRB, park2021novel} and non-saturating logistic loss with $R_{1}$ regularization~\cite{mescheder2018training, karras2020analyzing, karras2019style, karras2021alias}. In detail, the hinge loss is defined as

\begin{eqnarray}
\label{eq:hinge_D}
    \lefteqn{L_D=E_{x\sim P_{data}(x)}[\max(0, 1-D(x))]} \nonumber\\
    & & { \qquad} + E_{z\sim P_{z(z)}}[max(0, 1+D(G(z)))],
\end{eqnarray}
\begin{eqnarray}
\label{eq:hinge_G}
    L_G=-E_{z\sim P_{z(z)}}[D(G(z))],
\end{eqnarray}
whereas the non-saturating logistic loss with $R_{1}$ regularization is formulated as

\begin{eqnarray}
\label{eq:R1_D}
      \lefteqn{L_D = E_{x\sim P_\textrm{data}(x)}[\log(1 + e^{-D(x)}) + \cfrac{\gamma}{2}\|\nabla D(x)\|^{2} ]}\nonumber\\
    & & {\qquad \qquad} + E_{z\sim P_{z}(z)}[\log(1 + e^{D(G(z))})],
\end{eqnarray}
\begin{eqnarray}
\label{eq:R1_G}
    L_G = E_{z\sim P_{z}(z)}[\log(1 + e^{-D(G(z))})],
\end{eqnarray}
where $\gamma$ is a hyper-parameter that constrains the gradient magnitude of the discriminator. 

With reference to the existing papers~\cite{park2021GRB, park2021novel, karras2019style, karras2020analyzing}, we took the different training strategies according to the image resolution. When training the CIFAR-10 and CIFAR-100 datasets, we used hinge loss as an objective function. In addition, we employed the Adam optimizer~\cite{kingma2014adam} and set the user parameters of Adam optimizer, \ie $\beta_1$ and $\beta_2$, to 0 and 0.9, respectively. The learning rate was set to 2e-4, and the discriminator was updated 5 times using different mini-batches when the generator was updated once. We set the batch size of the discriminator as 64 and trained the generator for 50k iterations. In our experiments, the generator was trained with a batch size twice as large as when training the discriminator. That means the generator and discriminator were trained with 128 and 64 batch sizes, respectively. 

On the other hand, when training the FFHQ, AFHQ, and LSUN datasets, we set StyleGAN2~\cite{karras2020analyzing} as our baseline model and followed its training strategy since it is the most popular generative model generating a high-resolution image with plausible quality. Specifically, we employed the non-saturating logistic loss in Eqs.~\ref{eq:R1_D} and~\ref{eq:R1_G} where $\gamma$ was set to one. In addition, we set the $\beta_1$ and $\beta_2$ in the Adam optimizer to 0 and 0.99, respectively. Both batch sizes of the discriminator and generator were set to 32. The networks were trained while the generator tried to generate 10,000k images on the FFHQ and LSUN datasets, and 50,000k images on the AFHQ dataset. We also employed the lazy regularization technique~\cite{karras2020analyzing} where $R_1$ regularization was performed only once every 16 mini-batches. The more detailed implementation and training skills can be found in~\cite{karras2020analyzing}. \newline

\noindent
\textbf{Network Architecture} The goal of this paper is to show the ability of the proposed FCB. Thus, for a fair comparison, we compared the performance between the existing methods and the proposed one in the same environment except for the block module in the generator. Toward this end, we set the baseline model and only replaced the residual block in the baseline model with the FCB. In detail, on the CIFAR-10 and CIFAR-100 datasets, we decided the model in~\cite{miyato2018spectral} as our baseline model, whereas StyleGAN2~\cite{karras2020analyzing} was set to our baseline model for training the FFHQ, AFHQ, and LSUN datasets. The detailed network architectures of each baseline model are described in Tables.~\ref{table1} and~\ref{table2}. We implemented the SNGAN baseline model using \textit{Tensorflow} and employed the official \textit{Pytorch} code when implementing the StyleGAN2 baseline model. All models were trained on a single RTX 3090 GPU.\newline 

\subsection{Experimental Results}
\label{sec4.2}

\noindent
\textbf{Evaluation Metrics} In this paper, we employed the most popular metrics called FID~\cite{heusel2017gans} and Precision/Recall~\cite{kynkaanniemi2019improved, sajjadi2018assessing}, which quantify how realistic the generated image is. The FID measures the Wasserstein-2 distance between the feature distributions of the real and generated images, which are obtained via the pre-trained Inception model~\cite{szegedy2016rethinking}. Precision and Recall quantify the percentage of generated images that are similar to training images and the percentage of training images that can be generated, respectively. In our experiments, we generated 50,000 samples and computed the metrics. \newline

\begin{figure*}
\centering
\includegraphics[width=0.8\linewidth]{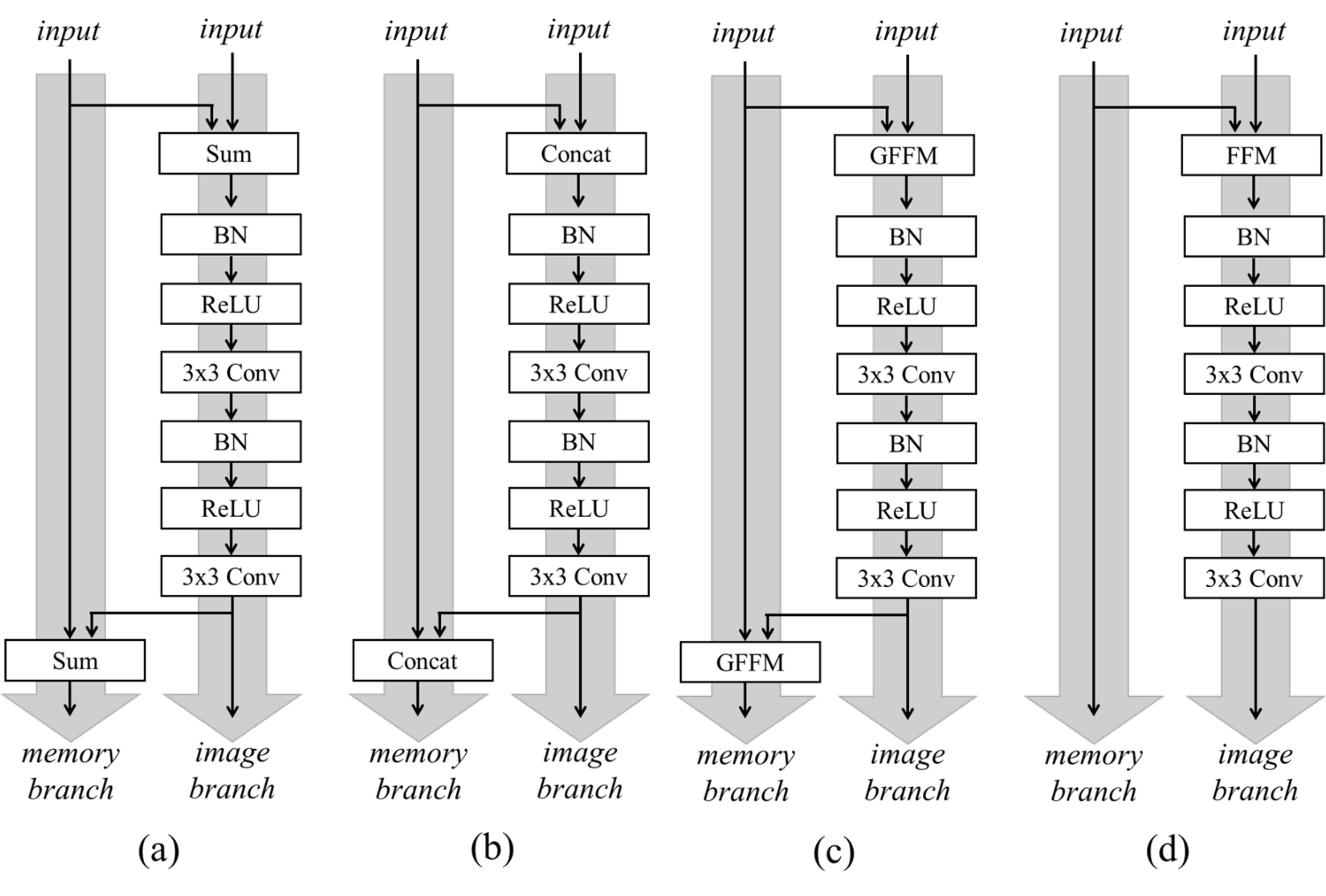}
\caption{Modified FCBs used for ablation studies. (a) $\textrm{FCB}_S$, (b) $\textrm{FCB}_C$, (c) $\textrm{FCB}_G$, (d) $\textrm{FCB}^\dagger$}
\label{fig4}
\vspace{-0cm}
\end{figure*}

\begin{table}[t]
\caption{Experimental results of ablation studies. We measured the GAN performance in terms of FID.}

\begin{center}
\begin{tabular}{c | c | c | c | c | c}
\hline
Dataset & & $\textrm{FCB}_S$ & $\textrm{FCB}_C$ & $\textrm{FCB}_G$ & $\textrm{FCB}^\dagger$ \\ 
\hline
\multirow{4}*{CIFAR-10} & trial 1 & 11.83 & 10.12 & 10.08 & 11.24 \\
 & trial 2 & 11.25 & 10.78 & 9.79 & 10.55 \\
 & trial 3 & 10.85 & 10.68 & 9.99 & 10.71 \\
\cline{2-6}
& Mean & 11.31 & 10.53 & 9.95 & 10.89 \\
\hline

\multirow{4}*{CIFAR-100} & trial 1 & 16.08 & 14.91 & 14.83 & 14.99 \\
 & trial 2 & 15.41 & 14.65 & 14.47 & 15.16 \\
 & trial 3 & 15.17 & 15.26 & 14.38 & 14.21 \\
\cline{2-6}
& Mean & 15.55 & 14.94 & 14.56 & 14.79 \\
\hline

\end{tabular}
\end{center}
\label{table3}
\end{table}

\noindent
\textbf{Quantitative Results} To highlight the superiority of the proposed FCB, we conducted preliminary experiments on the image generation task using the CIFAR-10 and CIFAR-100 datasets. In our experiments, we trained the network three times from scratch to clearly show that the performance gain was not due to the lucky weight initialization. Table~\ref{table2} summarizes the experimental results and the bold numbers indicate the best performance in each experiment. As shown in Table~\ref{table2}, on both CIFAR-10 and CIFAR-100 datasets, the proposed method consistently achieves better performance than its counterparts by a large margin in all trials. For instance, on the CIFAR-10 dataset, the FCB achieves the FID of 9.46, which is about 27.17\% better than the standard residual block (SNGAN). Moreover, the FCB exhibits about 11.34\% and 8.86\% better performance than the recent studies, \ie GRB and ABGAN. Based on these results, we concluded that the proposed method is effective for the generator compared to the conventional block modules. 

Furthermore, we investigated the ability of the proposed FFM by conducting ablation studies. To this end, we designed three different FCBs which use the summation, concatenation, and GFFM~\cite{park2021novel} instead of the FFM, respectively (we only modified the FFM part in the FCB with those operations). For better representation, we will refer to the FCB with the summation, concatenation, and GFFM as $\textrm{FCB}_S$, $\textrm{FCB}_C$, and $\textrm{FCB}_G$, respectively. The detailed architectures of each module are depicted in Fig.~\ref{fig4}. Table~\ref{table3} presents the experimental results. Here, we can interpret these results as follows: the $\textrm{FCB}_S$ and $\textrm{FCB}_C$ show better performances than SNGAN and BigGAN, which reveals that the FCB is more appropriate for G than the traditional residual block even using simple mathematical operations. In particular, $\textrm{FCB}_S$ uses the same number of network parameters as the residual block but shows much better performance; the only difference between the $\textrm{FCB}_S$ and the residual block is the existence of the additional branch in the FCB, \ie the memory branch. That means, by simply adding the simple memory branch, we can improve the GAN performance about 12.93\% and 10.43\% on the CIFAR-10 and CIFAR-100 datasets, respectively. In fact, the ABGAN also contains the auxiliary branch which looks similar to the memory branch of the FCB. Compared with the ABGAN, the $\textrm{FCB}_S$ and $\textrm{FCB}_C$ exhibit weak performance. That means even if we use the additional branch, the results would change on how we combine the features in the image branch with the additional ones. To make it more reliable, we conducted additional experiments that combine two branches using the existing feature fusion module called GFFM~\cite{park2021novel}, \ie $\textrm{FCB}_G$ in Fig.~\ref{fig4}(c). As reported in Table~\ref{table3}, $\textrm{FCB}_G$ exhibits better performance than $\textrm{FCB}_S$ and $\textrm{FCB}_C$, but shows worse performance compared with the FCB. That means, the proposed FFM is more compatible with the FCB than the existing module in~\cite{park2021novel}. Based on these observations, we confirmed that the proposed FFM has an important role in the FCB. 

On the other hand, we checked how effects the memory branch to the image branch. Toward this end, we removed the FFM in the memory branch and measured the performance ($\textrm{FCB}^\dagger$ in Fig.~\ref{fig4}(d)). That means the memory branch affects the image branch, but not vice versa; the memory branch does not be updated. Note that we do not conduct an ablation study about removing the FFM in the image branch, since the image branch cannot receive any information if that FFM does not exist. As shown in Table~\ref{table3}, even not updating the memory branch, $\textrm{FCB}^\dagger$ achieves better performance than the conventional methods having a single branch, \ie SNGAN, BigGAN, and GRB. That means the memory branch is an important key for improving the GAN performance. Although $\textrm{FCB}^\dagger$ shows better performance than the single branch generator, it exhibits poor performance compared with the ABGAN and FCB which have two branches. These results indicate that the joint update between memory and image branches is needed to enhance the \textit{G} ability.\newline

\begin{table*}[t]
\caption{Comparison of GAN performances using the StyleGAN2 baseline model. We evaluated the GAN performance in terms of FID, and P \& R.}
\begin{center}
\begin{tabular}{c | c  c  c | c  c  c | c  c  c | c  c  c}
\hline \hline

\multirow{2}*{Method} & \multicolumn{3}{c|}{FFHQ} & \multicolumn{3}{c|}{AFHQ} & \multicolumn{3}{c|}{LSUN Car} & \multicolumn{3}{c}{LSUN Horse} \\
\cline{2-13} 
 & FID$\downarrow$ & P$\uparrow$ & R$\uparrow$ & FID$\downarrow$ & P$\uparrow$ & R$\uparrow$ & FID$\downarrow$ & P$\uparrow$ & R$\uparrow$ & FID$\downarrow$ & P$\uparrow$ & R$\uparrow$ \\
\hline
StyleGAN2 (Residual block)~\cite{karras2020analyzing}  & 4.89 & 0.69 & 0.39 & 7.62 & 0.78 & 0.27 & 5.39 & 0.63 & 0.32 & 4.80 & 0.62 & 0.32 \\
StyleGAN2 with GRB~\cite{park2021GRB} & 3.92 & 0.67 & 0.47 & 7.55 & 0.76 & 0.28 & 5.37 & 0.61 & 0.36 & 4.30 & 0.60 & 0.39 \\
StyleGAN2 with ABGAN~\cite{park2021novel} & 3.96 & 0.65 & 0.48 & 7.62 & 0.78 & 0.26 & 5.26 & 0.60 & 0.40 & 4.20 & 0.62 & 0.38 \\
\textbf{StyleGAN2 with FCB (Proposed)} & 3.72 & 0.66 & 0.48 & 6.75 & 0.76 & 0.28 & 4.63 & 0.62 & 0.37 & 4.45 & 0.6 & 0.37 \\

\hline \hline
\end{tabular}
\end{center}
\label{table4}
\end{table*}

\begin{figure*}
\centering
\includegraphics[width=0.95\linewidth]{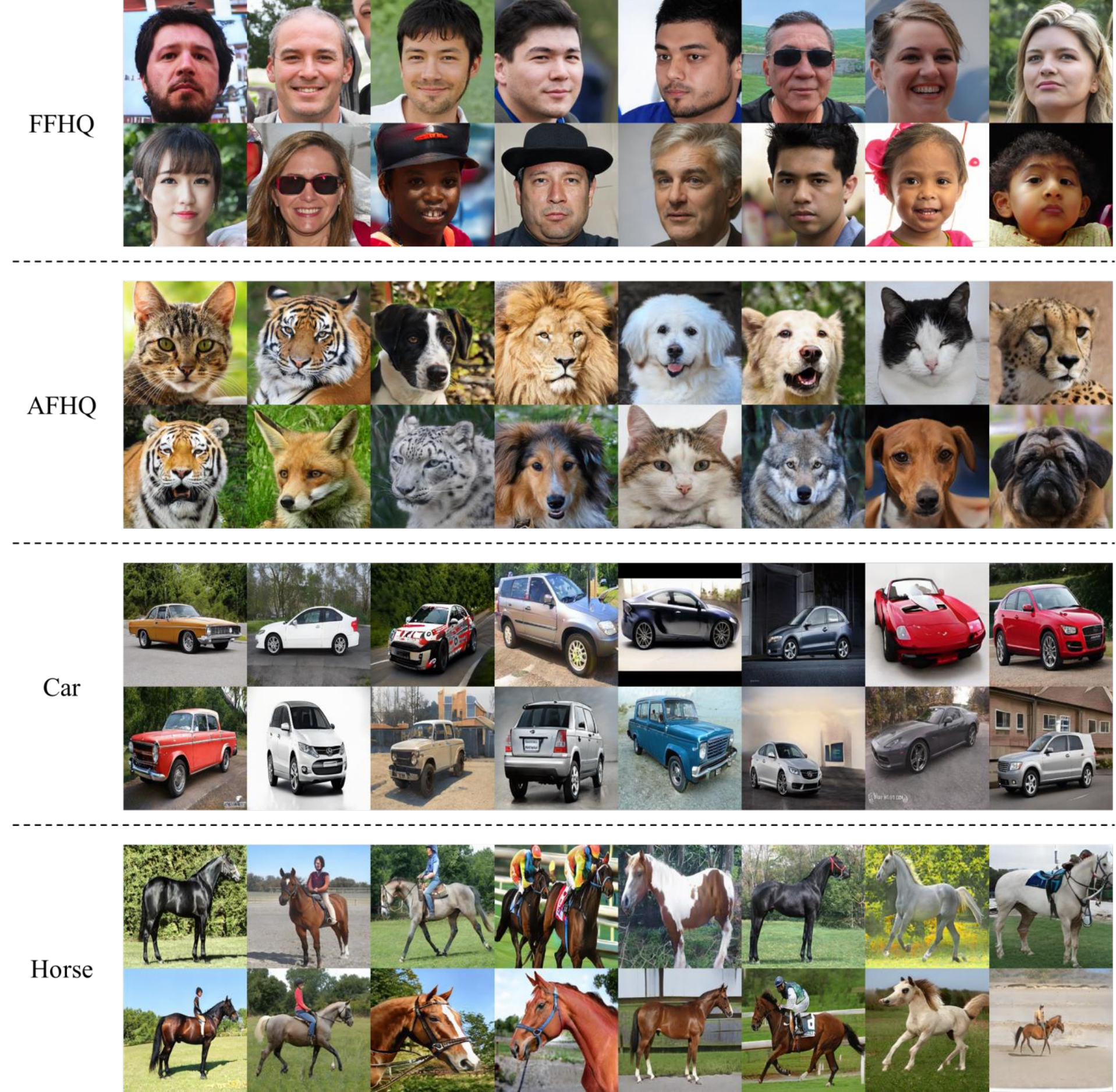}
\caption{Samples of the generated images using the proposed method on FFHQ, AFHQ, LSUN-Car, and LSUN-Horse datasets.}
\label{fig5}
\vspace{-0cm}
\end{figure*}

In addition, we trained our networks on the challenging dataset, \ie, the FFHQ, AFHQ, and LSUN datasets. Table~\ref{table4} presents the experimental results. From the perspective of FID, the proposed method substantially improves StyleGAN2 performance by a clear margin. These results strongly prove the effectiveness of the proposed FCB. Moreover, compared with the GRB and ABGAN, the proposed method further introduce significant gains in terms of FID, achieving impressive performance on various datasets. Namely, we expect that the proposed method could be compatible with various image generation tasks. In terms of Precision and Recall, compared with StylgeGAN2, the proposed method shows similar Precision scores but clear gains are observed in Recall across all datasets. That means the proposed method exhibits better  synthesis diversity than StyleGAN2. In fact, the proposed method shows similar Precision and Recall scores to the GRB and ABGAN. That means the GRB and ABGAN are also useful to generate diverse images like the proposed method. However, as mentioned above, the proposed method shows better performance in terms of FID than the GRB and ABGAN. Thus, we believe that the proposed method exhibits outstanding performance compared with the conventional methods. \newline

\noindent
\textbf{Qualitative Results} Fig.~\ref{fig5} shows the samples of the generated images. We agree that it is reasonable to compare the generated image quality of the conventional and the proposed methods. However, since random noise is used as the input to generate the images, it is difficult to fairly compare the image quality of conventional and proposed methods. Therefore, following the previous papers~\cite{brock2018large, yeo2021simple, park2022effective, park2022generative}, this paper only presents the generated images of the proposed method. As illustrated in Fig.~\ref{fig5}, our model can synthesize diverse and visually plausible images with high resolution. \newline 


\section{Conclusion}
\label{sec5}
We have presented a novel generative model having an information feedback system that jointly updates the memory and image branches. Toward this end, we have introduced a deep learning module called FCB that interchanges the information in the latent and feature spaces via the FFM. We have demonstrated the effectiveness of the proposed method by showing extensive experimental results. In addition, we have further investigated the validity of the proposed method in its broad aspects through ablation studies. Therefore, we believe that the proposed method is useful for various GAN-based image generation tasks.

In fact, despite the significant improvements, the proposed method still faces one confusion, \ie hardware efficiency. Since a single FCB block contains the two FFMs with additional convolutional layers, it needs more computational costs than existing methods; the proposed method requires slightly more GPU memory and computational time. However, as proved in Table.~\ref{table3}, the proposed method exhibits better performance than conventional residual block, even using simple mathematical operations instead of the FFM. Thus, in our future work, we will further investigate the simple but effective feature fusion module which could alleviate the computational complexity of the FCB. In addition, we believe that there could be an optimal feature fusion module for the FCB; there could be another module to achieve better performance than the proposed FFM. Thus, we also plan to further investigate the feature blending technique in our future study.

\section*{Declaration of Competing Interest}
The authors declare no conflict of interest.

\printcredits

\bibliographystyle{cas-model2-names}

\bibliography{egbib.bib}

\begin{thebibliography}{48}
\expandafter\ifx\csname natexlab\endcsname\relax\def\natexlab#1{#1}\fi
\providecommand{\url}[1]{\texttt{#1}}
\providecommand{\href}[2]{#2}
\providecommand{\path}[1]{#1}
\providecommand{\DOIprefix}{doi:}
\providecommand{\ArXivprefix}{arXiv:}
\providecommand{\URLprefix}{URL: }
\providecommand{\Pubmedprefix}{pmid:}
\providecommand{\doi}[1]{\href{http://dx.doi.org/#1}{\path{#1}}}
\providecommand{\Pubmed}[1]{\href{pmid:#1}{\path{#1}}}
\providecommand{\bibinfo}[2]{#2}
\ifx\xfnm\relax \def\xfnm[#1]{\unskip,\space#1}\fi
\bibitem[{Arjovsky et~al.(2017)Arjovsky, Chintala and
  Bottou}]{arjovsky2017wasserstein}
\bibinfo{author}{Arjovsky, M.}, \bibinfo{author}{Chintala, S.},
  \bibinfo{author}{Bottou, L.}, \bibinfo{year}{2017}.
\newblock \bibinfo{title}{Wasserstein generative adversarial networks}, in:
  \bibinfo{booktitle}{International conference on machine learning},
  \bibinfo{organization}{PMLR}. pp. \bibinfo{pages}{214--223}.
\bibitem[{Bai et~al.(2022)Bai, Yang, Xu, Liu, Yang and Shen}]{bai2022glead}
\bibinfo{author}{Bai, Q.}, \bibinfo{author}{Yang, C.}, \bibinfo{author}{Xu,
  Y.}, \bibinfo{author}{Liu, X.}, \bibinfo{author}{Yang, Y.},
  \bibinfo{author}{Shen, Y.}, \bibinfo{year}{2022}.
\newblock \bibinfo{title}{Glead: Improving gans with a generator-leading task}.
\newblock \bibinfo{journal}{arXiv preprint arXiv:2212.03752} .
\bibitem[{Brock et~al.(2018)Brock, Donahue and Simonyan}]{brock2018large}
\bibinfo{author}{Brock, A.}, \bibinfo{author}{Donahue, J.},
  \bibinfo{author}{Simonyan, K.}, \bibinfo{year}{2018}.
\newblock \bibinfo{title}{Large scale gan training for high fidelity natural
  image synthesis}.
\newblock \bibinfo{journal}{arXiv preprint arXiv:1809.11096} .
\bibitem[{Chen et~al.(2019)Chen, Zhai, Ritter, Lucic and
  Houlsby}]{chen2019self}
\bibinfo{author}{Chen, T.}, \bibinfo{author}{Zhai, X.},
  \bibinfo{author}{Ritter, M.}, \bibinfo{author}{Lucic, M.},
  \bibinfo{author}{Houlsby, N.}, \bibinfo{year}{2019}.
\newblock \bibinfo{title}{Self-supervised gans via auxiliary rotation loss},
  in: \bibinfo{booktitle}{Proceedings of the IEEE Conference on Computer Vision
  and Pattern Recognition}, pp. \bibinfo{pages}{12154--12163}.
\bibitem[{Choi et~al.(2018)Choi, Choi, Kim, Ha, Kim and Choo}]{choi2018stargan}
\bibinfo{author}{Choi, Y.}, \bibinfo{author}{Choi, M.}, \bibinfo{author}{Kim,
  M.}, \bibinfo{author}{Ha, J.W.}, \bibinfo{author}{Kim, S.},
  \bibinfo{author}{Choo, J.}, \bibinfo{year}{2018}.
\newblock \bibinfo{title}{Stargan: Unified generative adversarial networks for
  multi-domain image-to-image translation}, in: \bibinfo{booktitle}{Proceedings
  of the IEEE/CVF conference on computer vision and pattern recognition}, pp.
  \bibinfo{pages}{8789--8797}.
\bibitem[{Choi et~al.(2020)Choi, Uh, Yoo and Ha}]{choi2020stargan}
\bibinfo{author}{Choi, Y.}, \bibinfo{author}{Uh, Y.}, \bibinfo{author}{Yoo,
  J.}, \bibinfo{author}{Ha, J.W.}, \bibinfo{year}{2020}.
\newblock \bibinfo{title}{Stargan v2: Diverse image synthesis for multiple
  domains}, in: \bibinfo{booktitle}{Proceedings of the IEEE/CVF Conference on
  Computer Vision and Pattern Recognition}, pp. \bibinfo{pages}{8188--8197}.
\bibitem[{Goodfellow et~al.(2014)Goodfellow, Pouget-Abadie, Mirza, Xu,
  Warde-Farley, Ozair, Courville and Bengio}]{goodfellow2014generative}
\bibinfo{author}{Goodfellow, I.}, \bibinfo{author}{Pouget-Abadie, J.},
  \bibinfo{author}{Mirza, M.}, \bibinfo{author}{Xu, B.},
  \bibinfo{author}{Warde-Farley, D.}, \bibinfo{author}{Ozair, S.},
  \bibinfo{author}{Courville, A.}, \bibinfo{author}{Bengio, Y.},
  \bibinfo{year}{2014}.
\newblock \bibinfo{title}{Generative adversarial nets}, in:
  \bibinfo{booktitle}{Advances in neural information processing systems}, pp.
  \bibinfo{pages}{2672--2680}.
\bibitem[{Gulrajani et~al.(2017)Gulrajani, Ahmed, Arjovsky, Dumoulin and
  Courville}]{gulrajani2017improved}
\bibinfo{author}{Gulrajani, I.}, \bibinfo{author}{Ahmed, F.},
  \bibinfo{author}{Arjovsky, M.}, \bibinfo{author}{Dumoulin, V.},
  \bibinfo{author}{Courville, A.C.}, \bibinfo{year}{2017}.
\newblock \bibinfo{title}{Improved training of wasserstein gans}.
\newblock \bibinfo{journal}{Advances in neural information processing systems}
  \bibinfo{volume}{30}.
\bibitem[{Heusel et~al.(2017)Heusel, Ramsauer, Unterthiner, Nessler and
  Hochreiter}]{heusel2017gans}
\bibinfo{author}{Heusel, M.}, \bibinfo{author}{Ramsauer, H.},
  \bibinfo{author}{Unterthiner, T.}, \bibinfo{author}{Nessler, B.},
  \bibinfo{author}{Hochreiter, S.}, \bibinfo{year}{2017}.
\newblock \bibinfo{title}{{GANs trained by a two time-scale update rule
  converge to a local Nash equilibrium}}, in: \bibinfo{booktitle}{Advances in
  Neural Information Processing Systems}, pp. \bibinfo{pages}{6626--6637}.
\bibitem[{Hong et~al.(2018)Hong, Yang, Choi and Lee}]{hong2018inferring}
\bibinfo{author}{Hong, S.}, \bibinfo{author}{Yang, D.}, \bibinfo{author}{Choi,
  J.}, \bibinfo{author}{Lee, H.}, \bibinfo{year}{2018}.
\newblock \bibinfo{title}{Inferring semantic layout for hierarchical
  text-to-image synthesis}, in: \bibinfo{booktitle}{Proceedings of the IEEE/CVF
  conference on computer vision and pattern recognition}, pp.
  \bibinfo{pages}{7986--7994}.
\bibitem[{Isola et~al.(2017)Isola, Zhu, Zhou and Efros}]{isola2017image}
\bibinfo{author}{Isola, P.}, \bibinfo{author}{Zhu, J.Y.},
  \bibinfo{author}{Zhou, T.}, \bibinfo{author}{Efros, A.A.},
  \bibinfo{year}{2017}.
\newblock \bibinfo{title}{Image-to-image translation with conditional
  adversarial networks}, in: \bibinfo{booktitle}{Proceedings of the IEEE/CVF
  conference on computer vision and pattern recognition}, pp.
  \bibinfo{pages}{1125--1134}.
\bibitem[{Karras et~al.(2021)Karras, Aittala, Laine, H{\"a}rk{\"o}nen,
  Hellsten, Lehtinen and Aila}]{karras2021alias}
\bibinfo{author}{Karras, T.}, \bibinfo{author}{Aittala, M.},
  \bibinfo{author}{Laine, S.}, \bibinfo{author}{H{\"a}rk{\"o}nen, E.},
  \bibinfo{author}{Hellsten, J.}, \bibinfo{author}{Lehtinen, J.},
  \bibinfo{author}{Aila, T.}, \bibinfo{year}{2021}.
\newblock \bibinfo{title}{Alias-free generative adversarial networks}.
\newblock \bibinfo{journal}{Advances in Neural Information Processing Systems}
  \bibinfo{volume}{34}.
\bibitem[{Karras et~al.(2019)Karras, Laine and Aila}]{karras2019style}
\bibinfo{author}{Karras, T.}, \bibinfo{author}{Laine, S.},
  \bibinfo{author}{Aila, T.}, \bibinfo{year}{2019}.
\newblock \bibinfo{title}{A style-based generator architecture for generative
  adversarial networks}, in: \bibinfo{booktitle}{Proceedings of the IEEE/CVF
  conference on computer vision and pattern recognition}, pp.
  \bibinfo{pages}{4401--4410}.
\bibitem[{Karras et~al.(2020)Karras, Laine, Aittala, Hellsten, Lehtinen and
  Aila}]{karras2020analyzing}
\bibinfo{author}{Karras, T.}, \bibinfo{author}{Laine, S.},
  \bibinfo{author}{Aittala, M.}, \bibinfo{author}{Hellsten, J.},
  \bibinfo{author}{Lehtinen, J.}, \bibinfo{author}{Aila, T.},
  \bibinfo{year}{2020}.
\newblock \bibinfo{title}{Analyzing and improving the image quality of
  stylegan}, in: \bibinfo{booktitle}{Proceedings of the IEEE/CVF conference on
  computer vision and pattern recognition}, pp. \bibinfo{pages}{8110--8119}.
\bibitem[{Kingma and Ba(2014)}]{kingma2014adam}
\bibinfo{author}{Kingma, D.P.}, \bibinfo{author}{Ba, J.}, \bibinfo{year}{2014}.
\newblock \bibinfo{title}{Adam: A method for stochastic optimization}.
\newblock \bibinfo{journal}{arXiv preprint arXiv:1412.6980} .
\bibitem[{Kodali et~al.(2017)Kodali, Abernethy, Hays and
  Kira}]{kodali2017convergence}
\bibinfo{author}{Kodali, N.}, \bibinfo{author}{Abernethy, J.},
  \bibinfo{author}{Hays, J.}, \bibinfo{author}{Kira, Z.}, \bibinfo{year}{2017}.
\newblock \bibinfo{title}{On convergence and stability of gans}.
\newblock \bibinfo{journal}{arXiv preprint arXiv:1705.07215} .
\bibitem[{Krizhevsky et~al.(2009)Krizhevsky, Hinton
  et~al.}]{krizhevsky2009learning}
\bibinfo{author}{Krizhevsky, A.}, \bibinfo{author}{Hinton, G.}, et~al.,
  \bibinfo{year}{2009}.
\newblock \bibinfo{title}{Learning multiple layers of features from tiny
  images} .
\bibitem[{Kynk{\"a}{\"a}nniemi et~al.(2019)Kynk{\"a}{\"a}nniemi, Karras, Laine,
  Lehtinen and Aila}]{kynkaanniemi2019improved}
\bibinfo{author}{Kynk{\"a}{\"a}nniemi, T.}, \bibinfo{author}{Karras, T.},
  \bibinfo{author}{Laine, S.}, \bibinfo{author}{Lehtinen, J.},
  \bibinfo{author}{Aila, T.}, \bibinfo{year}{2019}.
\newblock \bibinfo{title}{Improved precision and recall metric for assessing
  generative models}.
\newblock \bibinfo{journal}{Advances in Neural Information Processing Systems}
  \bibinfo{volume}{32}.
\bibitem[{Lin et~al.(2021)Lin, Sekar and Fanti}]{lin2021spectral}
\bibinfo{author}{Lin, Z.}, \bibinfo{author}{Sekar, V.}, \bibinfo{author}{Fanti,
  G.}, \bibinfo{year}{2021}.
\newblock \bibinfo{title}{Why spectral normalization stabilizes gans: Analysis
  and improvements}.
\newblock \bibinfo{journal}{Advances in neural information processing systems}
  \bibinfo{volume}{34}.
\bibitem[{Mao et~al.(2017)Mao, Li, Xie, Lau, Wang and
  Paul~Smolley}]{mao2017least}
\bibinfo{author}{Mao, X.}, \bibinfo{author}{Li, Q.}, \bibinfo{author}{Xie, H.},
  \bibinfo{author}{Lau, R.Y.}, \bibinfo{author}{Wang, Z.},
  \bibinfo{author}{Paul~Smolley, S.}, \bibinfo{year}{2017}.
\newblock \bibinfo{title}{Least squares generative adversarial networks}, in:
  \bibinfo{booktitle}{Proceedings of the IEEE international conference on
  computer vision}, pp. \bibinfo{pages}{2794--2802}.
\bibitem[{Mescheder et~al.(2018)Mescheder, Geiger and
  Nowozin}]{mescheder2018training}
\bibinfo{author}{Mescheder, L.}, \bibinfo{author}{Geiger, A.},
  \bibinfo{author}{Nowozin, S.}, \bibinfo{year}{2018}.
\newblock \bibinfo{title}{Which training methods for gans do actually
  converge?}, in: \bibinfo{booktitle}{International conference on machine
  learning}, \bibinfo{organization}{PMLR}. pp. \bibinfo{pages}{3481--3490}.
\bibitem[{Miyato et~al.(2018)Miyato, Kataoka, Koyama and
  Yoshida}]{miyato2018spectral}
\bibinfo{author}{Miyato, T.}, \bibinfo{author}{Kataoka, T.},
  \bibinfo{author}{Koyama, M.}, \bibinfo{author}{Yoshida, Y.},
  \bibinfo{year}{2018}.
\newblock \bibinfo{title}{Spectral normalization for generative adversarial
  networks}.
\newblock \bibinfo{journal}{arXiv preprint arXiv:1802.05957} .
\bibitem[{Miyato and Koyama(2018)}]{miyato2018cgans}
\bibinfo{author}{Miyato, T.}, \bibinfo{author}{Koyama, M.},
  \bibinfo{year}{2018}.
\newblock \bibinfo{title}{cgans with projection discriminator}.
\newblock \bibinfo{journal}{arXiv preprint arXiv:1802.05637} .
\bibitem[{Park and Shin(2021a)}]{park2021GRB}
\bibinfo{author}{Park, S.}, \bibinfo{author}{Shin, Y.G.},
  \bibinfo{year}{2021}a.
\newblock \bibinfo{title}{Generative residual block for image generation}.
\newblock \bibinfo{journal}{Applied Intelligence} , \bibinfo{pages}{1--10}.
\bibitem[{Park and Shin(2021b)}]{park2021novel}
\bibinfo{author}{Park, S.}, \bibinfo{author}{Shin, Y.G.},
  \bibinfo{year}{2021}b.
\newblock \bibinfo{title}{A novel generator with auxiliary branch for improving
  gan performance}.
\newblock \bibinfo{journal}{arXiv preprint arXiv:2112.14968} .
\bibitem[{Park and Shin(2022)}]{park2022generative}
\bibinfo{author}{Park, S.}, \bibinfo{author}{Shin, Y.G.}, \bibinfo{year}{2022}.
\newblock \bibinfo{title}{Generative convolution layer for image generation}.
\newblock \bibinfo{journal}{Neural Networks} .
\bibitem[{Park et~al.(2022a)Park, Yeo and Shin}]{park2022pconv}
\bibinfo{author}{Park, S.}, \bibinfo{author}{Yeo, Y.J.}, \bibinfo{author}{Shin,
  Y.G.}, \bibinfo{year}{2022}a.
\newblock \bibinfo{title}{Pconv: simple yet effective convolutional layer for
  generative adversarial network}.
\newblock \bibinfo{journal}{Neural Computing and Applications}
  \bibinfo{volume}{34}, \bibinfo{pages}{7113--7124}.
\bibitem[{Park et~al.(2022b)Park, Yoo and Shin}]{park2022effective}
\bibinfo{author}{Park, S.}, \bibinfo{author}{Yoo, C.H.}, \bibinfo{author}{Shin,
  Y.G.}, \bibinfo{year}{2022}b.
\newblock \bibinfo{title}{Effective shortcut technique for generative
  adversarial networks}.
\newblock \bibinfo{journal}{Applied Intelligence} , \bibinfo{pages}{1--13}.
\bibitem[{Reed et~al.(2016)Reed, Akata, Yan, Logeswaran, Schiele and
  Lee}]{reed2016generative}
\bibinfo{author}{Reed, S.}, \bibinfo{author}{Akata, Z.}, \bibinfo{author}{Yan,
  X.}, \bibinfo{author}{Logeswaran, L.}, \bibinfo{author}{Schiele, B.},
  \bibinfo{author}{Lee, H.}, \bibinfo{year}{2016}.
\newblock \bibinfo{title}{Generative adversarial text to image synthesis}.
\newblock \bibinfo{journal}{arXiv preprint arXiv:1605.05396} .
\bibitem[{Ross and Doshi-Velez(2018)}]{ross2018improving}
\bibinfo{author}{Ross, A.}, \bibinfo{author}{Doshi-Velez, F.},
  \bibinfo{year}{2018}.
\newblock \bibinfo{title}{Improving the adversarial robustness and
  interpretability of deep neural networks by regularizing their input
  gradients}, in: \bibinfo{booktitle}{Proceedings of the AAAI Conference on
  Artificial Intelligence}, pp. \bibinfo{pages}{1660--1669}.
\bibitem[{Roth et~al.(2017)Roth, Lucchi, Nowozin and
  Hofmann}]{roth2017stabilizing}
\bibinfo{author}{Roth, K.}, \bibinfo{author}{Lucchi, A.},
  \bibinfo{author}{Nowozin, S.}, \bibinfo{author}{Hofmann, T.},
  \bibinfo{year}{2017}.
\newblock \bibinfo{title}{Stabilizing training of generative adversarial
  networks through regularization}, in: \bibinfo{booktitle}{Advances in neural
  information processing systems}, pp. \bibinfo{pages}{2018--2028}.
\bibitem[{Sagong et~al.(2019)Sagong, Shin, Kim, Park and Ko}]{sagong2019pepsi}
\bibinfo{author}{Sagong, M.C.}, \bibinfo{author}{Shin, Y.G.},
  \bibinfo{author}{Kim, S.W.}, \bibinfo{author}{Park, S.}, \bibinfo{author}{Ko,
  S.J.}, \bibinfo{year}{2019}.
\newblock \bibinfo{title}{Pepsi: Fast image inpainting with parallel decoding
  network}, in: \bibinfo{booktitle}{Proceedings of the IEEE/CVF conference on
  computer vision and pattern recognition}, pp. \bibinfo{pages}{11360--11368}.
\bibitem[{Sagong et~al.(2022)Sagong, Yeo, Shin and Ko}]{sagong2022conditional}
\bibinfo{author}{Sagong, M.C.}, \bibinfo{author}{Yeo, Y.J.},
  \bibinfo{author}{Shin, Y.G.}, \bibinfo{author}{Ko, S.J.},
  \bibinfo{year}{2022}.
\newblock \bibinfo{title}{Conditional convolution projecting latent vectors on
  condition-specific space}.
\newblock \bibinfo{journal}{IEEE Transactions on Neural Networks and Learning
  Systems} .
\bibitem[{Sajjadi et~al.(2018)Sajjadi, Bachem, Lucic, Bousquet and
  Gelly}]{sajjadi2018assessing}
\bibinfo{author}{Sajjadi, M.S.}, \bibinfo{author}{Bachem, O.},
  \bibinfo{author}{Lucic, M.}, \bibinfo{author}{Bousquet, O.},
  \bibinfo{author}{Gelly, S.}, \bibinfo{year}{2018}.
\newblock \bibinfo{title}{Assessing generative models via precision and
  recall}.
\newblock \bibinfo{journal}{Advances in neural information processing systems}
  \bibinfo{volume}{31}.
\bibitem[{Shin et~al.(2020)Shin, Sagong, Yeo, Kim and Ko}]{shin2020pepsi++}
\bibinfo{author}{Shin, Y.G.}, \bibinfo{author}{Sagong, M.C.},
  \bibinfo{author}{Yeo, Y.J.}, \bibinfo{author}{Kim, S.W.},
  \bibinfo{author}{Ko, S.J.}, \bibinfo{year}{2020}.
\newblock \bibinfo{title}{Pepsi++: fast and lightweight network for image
  inpainting}.
\newblock \bibinfo{journal}{IEEE Transactions on Neural Networks and Learning
  Systems} .
\bibitem[{Szegedy et~al.(2016)Szegedy, Vanhoucke, Ioffe, Shlens and
  Wojna}]{szegedy2016rethinking}
\bibinfo{author}{Szegedy, C.}, \bibinfo{author}{Vanhoucke, V.},
  \bibinfo{author}{Ioffe, S.}, \bibinfo{author}{Shlens, J.},
  \bibinfo{author}{Wojna, Z.}, \bibinfo{year}{2016}.
\newblock \bibinfo{title}{Rethinking the inception architecture for computer
  vision}, in: \bibinfo{booktitle}{Proceedings of the IEEE conference on
  computer vision and pattern recognition}, pp. \bibinfo{pages}{2818--2826}.
\bibitem[{Wang et~al.(2021)Wang, Guo, Liu and Wei}]{wang2021up}
\bibinfo{author}{Wang, Y.}, \bibinfo{author}{Guo, X.}, \bibinfo{author}{Liu,
  P.}, \bibinfo{author}{Wei, B.}, \bibinfo{year}{2021}.
\newblock \bibinfo{title}{Up and down residual blocks for convolutional
  generative adversarial networks}.
\newblock \bibinfo{journal}{IEEE Access} \bibinfo{volume}{9},
  \bibinfo{pages}{26051--26058}.
\bibitem[{Wei et~al.(2018)Wei, Gong, Liu, Lu and Wang}]{wei2018improving}
\bibinfo{author}{Wei, X.}, \bibinfo{author}{Gong, B.}, \bibinfo{author}{Liu,
  Z.}, \bibinfo{author}{Lu, W.}, \bibinfo{author}{Wang, L.},
  \bibinfo{year}{2018}.
\newblock \bibinfo{title}{Improving the improved training of wasserstein gans:
  A consistency term and its dual effect}.
\newblock \bibinfo{journal}{arXiv preprint arXiv:1803.01541} .
\bibitem[{Wu et~al.(2019)Wu, Zhao, Chen, Xu, Wang, Zhang, Sun and
  Zhou}]{wu2019generalization}
\bibinfo{author}{Wu, B.}, \bibinfo{author}{Zhao, S.}, \bibinfo{author}{Chen,
  C.}, \bibinfo{author}{Xu, H.}, \bibinfo{author}{Wang, L.},
  \bibinfo{author}{Zhang, X.}, \bibinfo{author}{Sun, G.},
  \bibinfo{author}{Zhou, J.}, \bibinfo{year}{2019}.
\newblock \bibinfo{title}{Generalization in generative adversarial networks: A
  novel perspective from privacy protection}.
\newblock \bibinfo{journal}{arXiv preprint arXiv:1908.07882} .
\bibitem[{Wu et~al.(2021)Wu, Shuai, Tam and Chiu}]{wu2021gradient}
\bibinfo{author}{Wu, Y.L.}, \bibinfo{author}{Shuai, H.H.},
  \bibinfo{author}{Tam, Z.R.}, \bibinfo{author}{Chiu, H.Y.},
  \bibinfo{year}{2021}.
\newblock \bibinfo{title}{Gradient normalization for generative adversarial
  networks}.
\newblock \bibinfo{journal}{arXiv preprint arXiv:2109.02235} .
\bibitem[{Yang et~al.(2022)Yang, Shen, Xu, Zhao, Dai and
  Zhou}]{yang2022improving}
\bibinfo{author}{Yang, C.}, \bibinfo{author}{Shen, Y.}, \bibinfo{author}{Xu,
  Y.}, \bibinfo{author}{Zhao, D.}, \bibinfo{author}{Dai, B.},
  \bibinfo{author}{Zhou, B.}, \bibinfo{year}{2022}.
\newblock \bibinfo{title}{Improving gans with a dynamic discriminator}.
\newblock \bibinfo{journal}{arXiv preprint arXiv:2209.09897} .
\bibitem[{Yeo et~al.(2022)Yeo, Sagong, Park, Ko and Shin}]{yeo2022image}
\bibinfo{author}{Yeo, Y.J.}, \bibinfo{author}{Sagong, M.C.},
  \bibinfo{author}{Park, S.}, \bibinfo{author}{Ko, S.J.},
  \bibinfo{author}{Shin, Y.G.}, \bibinfo{year}{2022}.
\newblock \bibinfo{title}{Image generation with self pixel-wise normalization}.
\newblock \bibinfo{journal}{arXiv preprint arXiv:2201.10725} .
\bibitem[{Yeo et~al.(2021)Yeo, Shin, Park and Ko}]{yeo2021simple}
\bibinfo{author}{Yeo, Y.J.}, \bibinfo{author}{Shin, Y.G.},
  \bibinfo{author}{Park, S.}, \bibinfo{author}{Ko, S.J.}, \bibinfo{year}{2021}.
\newblock \bibinfo{title}{Simple yet effective way for improving the
  performance of gan.}
\newblock \bibinfo{journal}{IEEE Transactions on Neural Networks and Learning
  Systems} .
\bibitem[{Yu et~al.(2015)Yu, Zhang, Song, Seff and Xiao}]{yu15lsun}
\bibinfo{author}{Yu, F.}, \bibinfo{author}{Zhang, Y.}, \bibinfo{author}{Song,
  S.}, \bibinfo{author}{Seff, A.}, \bibinfo{author}{Xiao, J.},
  \bibinfo{year}{2015}.
\newblock \bibinfo{title}{Lsun: Construction of a large-scale image dataset
  using deep learning with humans in the loop}.
\newblock \bibinfo{journal}{arXiv preprint arXiv:1506.03365} .
\bibitem[{Yu et~al.(2018)Yu, Lin, Yang, Shen, Lu and Huang}]{yu2018free}
\bibinfo{author}{Yu, J.}, \bibinfo{author}{Lin, Z.}, \bibinfo{author}{Yang,
  J.}, \bibinfo{author}{Shen, X.}, \bibinfo{author}{Lu, X.},
  \bibinfo{author}{Huang, T.S.}, \bibinfo{year}{2018}.
\newblock \bibinfo{title}{Free-form image inpainting with gated convolution}.
\newblock \bibinfo{journal}{arXiv preprint arXiv:1806.03589} .
\bibitem[{Zhang et~al.(2019)Zhang, Zhang, Odena and Lee}]{zhang2019consistency}
\bibinfo{author}{Zhang, H.}, \bibinfo{author}{Zhang, Z.},
  \bibinfo{author}{Odena, A.}, \bibinfo{author}{Lee, H.}, \bibinfo{year}{2019}.
\newblock \bibinfo{title}{Consistency regularization for generative adversarial
  networks}.
\newblock \bibinfo{journal}{arXiv preprint arXiv:1910.12027} .
\bibitem[{Zhao et~al.(2020)Zhao, Singh, Lee, Zhang, Odena and
  Zhang}]{zhao2020improved}
\bibinfo{author}{Zhao, Z.}, \bibinfo{author}{Singh, S.}, \bibinfo{author}{Lee,
  H.}, \bibinfo{author}{Zhang, Z.}, \bibinfo{author}{Odena, A.},
  \bibinfo{author}{Zhang, H.}, \bibinfo{year}{2020}.
\newblock \bibinfo{title}{Improved consistency regularization for gans}.
\newblock \bibinfo{journal}{arXiv preprint arXiv:2002.04724} .
\bibitem[{Zhu et~al.(2017)Zhu, Park, Isola and Efros}]{zhu2017unpaired}
\bibinfo{author}{Zhu, J.Y.}, \bibinfo{author}{Park, T.},
  \bibinfo{author}{Isola, P.}, \bibinfo{author}{Efros, A.A.},
  \bibinfo{year}{2017}.
\newblock \bibinfo{title}{Unpaired image-to-image translation using
  cycle-consistent adversarial networks}, in: \bibinfo{booktitle}{Proceedings
  of the IEEE international conference on computer vision}, pp.
  \bibinfo{pages}{2223--2232}.

\end{thebibliography}

\bio{}
\endbio

\endbio

\end{document}